\documentclass[runningheads]{llncs}
\usepackage[T1]{fontenc}
\usepackage{graphicx,verbatim}
\usepackage{amssymb}
\usepackage{subcaption}
\usepackage{multirow}

\newcommand{\w}{\textit{w/} }
\newcommand{\wo}{\textit{w/o} }

\usepackage[table]{xcolor}
\newcommand{\heatRet}[1]{%
  \ifdim #1 pt < 6 pt   \cellcolor{white}#1%
  \else\ifdim #1 pt < 12 pt \cellcolor{red!5}#1%
  \else\ifdim #1 pt < 18 pt \cellcolor{red!10}#1%
  \else\ifdim #1 pt < 24 pt \cellcolor{red!15}#1%
  \else\ifdim #1 pt < 30 pt \cellcolor{red!20}#1%
  \else\ifdim #1 pt < 36 pt \cellcolor{red!25}#1%
  \else\ifdim #1 pt < 42 pt \cellcolor{red!30}#1%
  \else\ifdim #1 pt < 48 pt \cellcolor{red!35}#1%
  \else\ifdim #1 pt < 54 pt \cellcolor{red!40}#1%
  \else\ifdim #1 pt < 60 pt \cellcolor{red!45}#1%
  \else\ifdim #1 pt < 66 pt \cellcolor{red!50}#1%
  \else \cellcolor{red!60}#1%
  \fi\fi\fi\fi\fi\fi\fi\fi\fi\fi\fi
}

\newcommand{\heatRo}[1]{%
  \ifdim #1 pt < 52.00 pt \cellcolor{white}#1%
  \else\ifdim #1 pt < 52.55 pt \cellcolor{red!3}#1%
  \else\ifdim #1 pt < 53.10 pt \cellcolor{red!6}#1%
  \else\ifdim #1 pt < 53.65 pt \cellcolor{red!9}#1%
  \else\ifdim #1 pt < 54.20 pt \cellcolor{red!12}#1%
  \else\ifdim #1 pt < 54.75 pt \cellcolor{red!15}#1%
  \else\ifdim #1 pt < 55.30 pt \cellcolor{red!18}#1%
  \else\ifdim #1 pt < 55.85 pt \cellcolor{red!21}#1%
  \else\ifdim #1 pt < 56.40 pt \cellcolor{red!24}#1%
  \else\ifdim #1 pt < 56.95 pt \cellcolor{red!27}#1%
  \else\ifdim #1 pt < 57.50 pt \cellcolor{red!30}#1%
  \else\ifdim #1 pt < 58.05 pt \cellcolor{red!33}#1%
  \else\ifdim #1 pt < 58.60 pt \cellcolor{red!36}#1%
  \else\ifdim #1 pt < 59.15 pt \cellcolor{red!39}#1%
  \else\ifdim #1 pt < 59.70 pt \cellcolor{red!42}#1%
  \else\ifdim #1 pt < 60.25 pt \cellcolor{red!45}#1%
  \else\ifdim #1 pt < 60.80 pt \cellcolor{red!48}#1%
  \else\ifdim #1 pt < 61.35 pt \cellcolor{red!51}#1%
  \else\ifdim #1 pt < 61.90 pt \cellcolor{red!54}#1%
  \else\ifdim #1 pt < 62.45 pt \cellcolor{red!57}#1%
  \else \cellcolor{red!60}#1%
  \fi\fi\fi\fi\fi\fi\fi\fi\fi\fi
  \fi\fi\fi\fi\fi\fi\fi\fi\fi\fi
}

\newcommand{\heatDet}[1]{%
  \ifdim #1 pt < 17.0 pt \cellcolor{white}#1%
  \else\ifdim #1 pt < 17.4 pt \cellcolor{red!3}#1%
  \else\ifdim #1 pt < 17.8 pt \cellcolor{red!6}#1%
  \else\ifdim #1 pt < 18.2 pt \cellcolor{red!9}#1%
  \else\ifdim #1 pt < 18.6 pt \cellcolor{red!12}#1%
  \else\ifdim #1 pt < 19.0 pt \cellcolor{red!15}#1%
  \else\ifdim #1 pt < 19.4 pt \cellcolor{red!18}#1%
  \else\ifdim #1 pt < 19.8 pt \cellcolor{red!21}#1%
  \else\ifdim #1 pt < 20.2 pt \cellcolor{red!24}#1%
  \else\ifdim #1 pt < 20.6 pt \cellcolor{red!27}#1%
  \else\ifdim #1 pt < 21.0 pt \cellcolor{red!30}#1%
  \else\ifdim #1 pt < 21.4 pt \cellcolor{red!33}#1%
  \else\ifdim #1 pt < 21.8 pt \cellcolor{red!36}#1%
  \else\ifdim #1 pt < 22.2 pt \cellcolor{red!39}#1%
  \else\ifdim #1 pt < 22.6 pt \cellcolor{red!42}#1%
  \else\ifdim #1 pt < 23.0 pt \cellcolor{red!45}#1%
  \else\ifdim #1 pt < 23.4 pt \cellcolor{red!48}#1%
  \else\ifdim #1 pt < 23.8 pt \cellcolor{red!51}#1%
  \else\ifdim #1 pt < 24.2 pt \cellcolor{red!54}#1%
  \else\ifdim #1 pt < 24.6 pt \cellcolor{red!57}#1%
  \else \cellcolor{red!60}#1%
  \fi\fi\fi\fi\fi\fi\fi\fi\fi\fi
  \fi\fi\fi\fi\fi\fi\fi\fi\fi\fi
}

\begin{document}
\title{LoFi: Location-Aware Fine-Grained Representation Learning for Chest X-ray}
%

\author{Myeongkyun~Kang\inst{1} \and Yanting~Yang\inst{1,2} \and Xiaoxiao~Li\inst{1,2}\textsuperscript{\dag}}  
\authorrunning{M. Kang et al.}
\institute{The University of British Columbia, Vancouver, BC V6T 1Z4, Canada\\
\email{xiaoxiao.li@ece.ubc.ca} \and
Vector Institute, Toronto, ON M5G 0C6, Canada}

\renewcommand{\thefootnote}{\dag}
\footnotetext{Corresponding author.}

\maketitle              
\begin{abstract}
Fine-grained representation learning is crucial for retrieval and phrase grounding in chest X-rays, where clinically relevant findings are often spatially confined. However, the lack of region-level supervision in contrastive models and the limited ability of large vision language models to capture fine-grained representations in external validation lead to suboptimal performance on these tasks. To address these limitations, we propose Location-aware Fine-grained representation learning (LoFi), which jointly optimizes sigmoid, captioning, and location-aware captioning losses using a lightweight large language model. The location-aware captioning loss enables region-level supervision through grounding and dense captioning objectives, thereby facilitating fine-grained representation learning. Building upon these representations, we integrate a fine-grained encoder into retrieval-based in-context learning to enhance chest X-ray grounding across diverse settings. Extensive experiments demonstrate that our method achieves superior retrieval and phrase grounding performance on MIMIC-CXR and PadChest-GR.
\end{abstract}
%
%
%


\section{Introduction}
Clinically relevant findings in medical imaging are often spatially confined rather than semantically distinct, necessitating fine-grained representation learning \cite{haghighi2024self}. Retrieval tasks identify relevant chest X-rays among similar candidates, requiring fine-grained discrimination \cite{zhang2025radir}, and phrase grounding tasks localize objects based on textual descriptions, demanding fine-grained cross-modal alignment \cite{de2025padchest}. However, medical contrastive models \cite{lai2024carzero,zhang2025radir} and large vision language models (LVLMs) \cite{bannur2024maira,deperrois2025radvlm} remain limited in their ability to capture subtle and fine-grained representations, leading to suboptimal performance on these tasks.

Contrastive chest X-ray models have been proposed to effectively leverage paired radiology reports for fine-grained, clinically relevant representation learning \cite{lai2024carzero,zhang2025radir}.
However, the absence of region-level supervision (e.g., bounding boxes) during training limits their ability to learn fine-grained representations; incorporating such supervision can alleviate this limitation.
Nevertheless, incorporating image-text-box triples in medical domains remains challenging due to the scarcity of datasets with region-level annotations.
Therefore, leveraging a lightweight large language model (LLM) \cite{team2025gemma} and a region-level pretrained encoder \cite{tschannen2025siglip} is desirable to address the challenges of medical image-text-box alignment.

A further challenge is that existing models, including LVLMs, remain limited in generalizing fine-grained representations under external validation \cite{li2025knowledge}.
To address this challenge, retrieval-based in-context learning (ICL) has been proposed to enable adaptation to new tasks by leveraging retrieved demonstrations at inference time \cite{tsimpoukelli2021multimodal,yang2022empirical}.
However, its performance largely depends on the quality of the retrieved demonstrations, underscoring the need for fine-grained representations and learning objectives that facilitate them.

We propose \textbf{Lo}cation-aware \textbf{Fi}ne-grained representation learning (LoFi), which jointly optimizes sigmoid, captioning, and location-aware captioning losses with a lightweight LLM \cite{team2025gemma} and a region-level pretrained encoder \cite{tschannen2025siglip}. The captioning loss facilitates learning from long-form text, while the location-aware captioning loss enables region-level supervision through grounding and dense captioning objectives, thereby enhancing fine-grained representation learning.
Building upon these representations, we integrate a fine-grained encoder into retrieval-based ICL to enhance chest X-ray grounding across diverse clinical settings.
We train our model on 208,949 radiology reports and 394,835 text-box pairs from MIMIC-CXR, and demonstrate the superiority of our fine-grained representations through retrieval tasks on MIMIC-CXR, as well as internal and external phrase grounding tasks on PadChest-GR.

In summary, the contributions are as follows:
\emph{(a)} We propose LoFi, which leverages a lightweight LLM to jointly optimize sigmoid, captioning, and location-aware captioning losses for fine-grained chest X-ray representation learning.
\emph{(b)} We integrate a fine-grained encoder into retrieval-based ICL, providing a practical solution for real-world chest X-ray grounding.
\emph{(c)} We demonstrate the effectiveness of our approach on retrieval and grounding tasks through comprehensive internal and external validation.
\begin{figure}[t]
    \centering
    \includegraphics[width=1.0\textwidth]{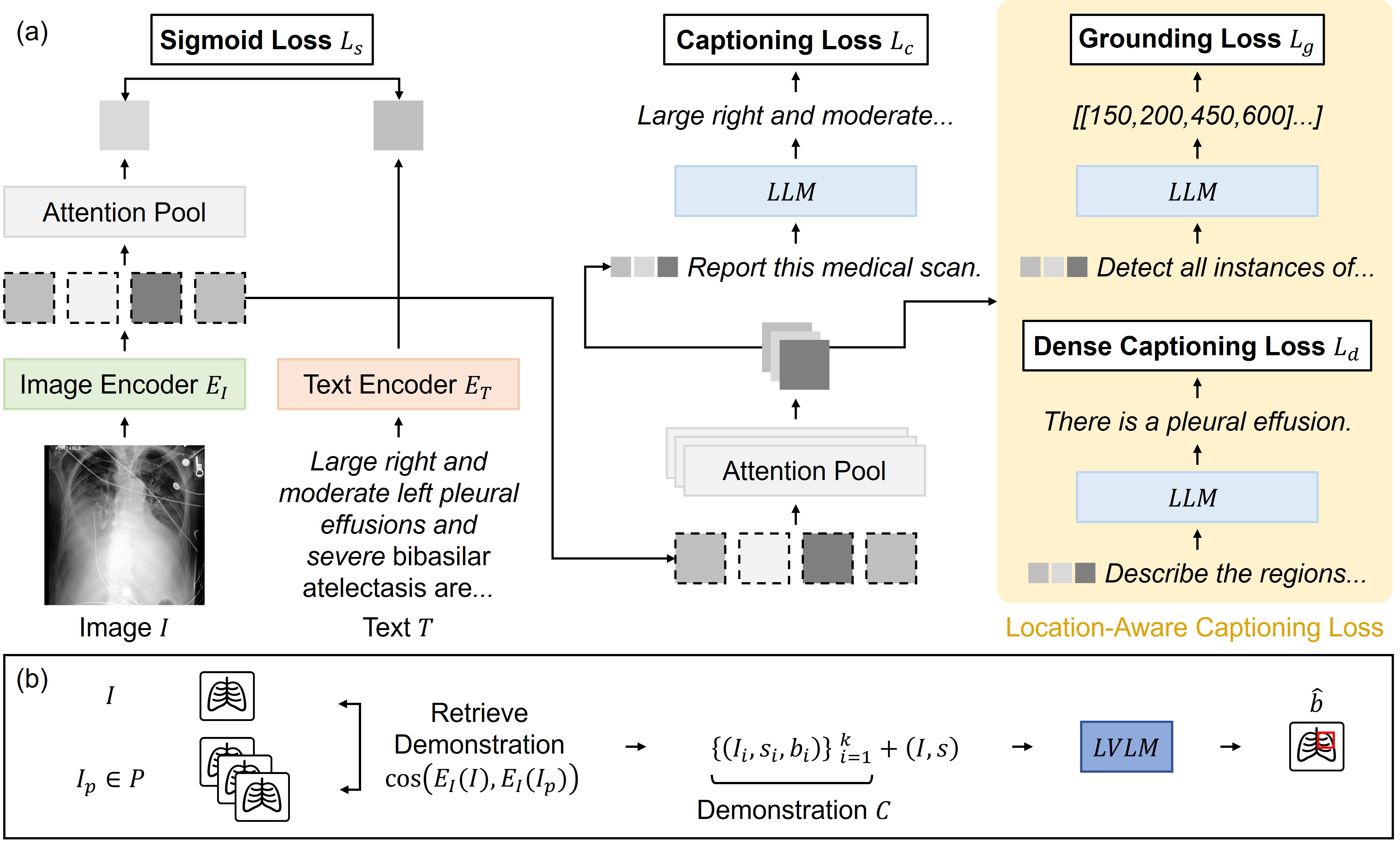}
    \caption{Illustration of (a) location-aware fine-grained representation learning frameworks and (b) a retrieval-based in-context learning pipeline.}
    \label{fig:method}
\end{figure}

\section{Method}

\subsubsection{Location-Aware Fine-Grained Representation Learning (LoFi)}
The proposed method is shown in Fig. \ref{fig:method}(a).
Given a dataset $\mathcal{D}$ of image-text-box triplets $(I,T,B)$, where box annotations may be optional, we aim to train an image encoder $E_I$ and a text encoder $E_T$, leveraging a lightweight LLM $LLM$.
We employ 64 attention pooling modules and a projection layer to align the $E_I$ outputs with the $LLM$ input dimension (notation omitted for simplicity).
We implement three objectives to facilitate learning from long-form text and region-level supervision: (a) a sigmoid loss, (b) a captioning loss, and (c) a location-aware captioning loss consisting of a grounding loss and a dense captioning loss.
We adopt a sigmoid loss \cite{tschannen2025siglip} as our contrastive objective, encouraging the model to assign higher similarity scores to matched pairs and thus learn representations that are well suited for retrieval.
Formally, the sigmoid loss is defined as
\begin{equation}
\mathcal{L}_s
= -\mathbb{E}_{(I,T)\sim \mathcal{D}}
\left[
\log \sigma\!\left(\delta(I,T)\, E_I(I)^\top E_T(T)\right)
\right],
\end{equation}
where $\sigma(\cdot)$ denotes the sigmoid, and $\delta(I,T)\in\{+1,-1\}$ is a sign indicator that equals $+1$ for matched positive pairs and $-1$ for in-batch negative pairs.
However, the sigmoid loss alone is limited in leveraging long-form text and region-level supervision of fine-grained object locations during training.
To address this issue, we adapt prior works \cite{tschannen2025siglip,wan2024locca} to our learning objective using a lightweight LLM.
We denote the conditional distribution induced by the $LLM$ as $p_{LLM}$, and define the autoregressive captioning loss as
\begin{equation}
\mathcal{L}_c = -\mathbb{E}_{(I,T)\sim \mathcal{D}}
\left[
\log p_{LLM}(T \mid E_I(I), \mathcal{P}_c)
\right],
\end{equation}
where $\mathcal{P}_c$ denotes a fixed prompt (e.g., \textit{``Report this medical scan.''}).
Next, to incorporate location information of fine-grained objects during training, we adapt the grounding objective and the dense captioning objective to our autoregressive loss.
Note that a text (e.g., a radiology report) $T$ consists of sentences, and we denote a sentence as $s \in T$.
We denote the set of bounding boxes associated with $s$ by $b \subseteq B$, where each box is represented by the coordinates of its top-left and bottom-right corners $(x_{\min}, y_{\min}, x_{\max}, y_{\max})$.
Formally, we define the grounding loss as
\begin{equation}
\mathcal{L}_g
=
-\mathbb{E}_{(I,T,B)\sim \mathcal{D}_B,\; s\sim T}
\left[
\log p_{LLM}(b \mid E_I(I), \mathcal{P}_g, s)
\right],
\end{equation}
where $\mathcal{D}_B \subseteq \mathcal{D}$ denotes samples with bounding boxes, and $\mathcal{P}_g$ denotes a fixed prompt (e.g., \textit{``Detect all instances of...''}).
For the dense captioning loss, we swap the prediction target from $b$ to $s$.
Formally, we define the dense captioning loss as
$
\mathcal{L}_d
=
-\mathbb{E}_{(I,T,B)\sim \mathcal{D}_B,\; s\sim T}
\left[
\log p_{LLM}(s \mid E_I(I), \mathcal{P}_d, b)
\right],
$
where $\mathcal{P}_d$ denotes a fixed prompt (e.g., \textit{``Describe the regions...''}).
In summary, the total loss is defined as $\mathcal{L}_t = \mathcal{L}_s + \lambda \cdot \mathcal{L}_{\tau}, \tau \in \{c, g, d\}$, where $\tau$ indicates the loss type (captioning, grounding, or dense captioning), and $\lambda$ is a hyperparameter for weighting the autoregressive loss (set to 5).
When $(I,T,B)$ are available, $\mathcal{L}_c$ and the location-aware captioning loss $\mathcal{L}_{g,d}$ contribute equally to the total loss.

\subsubsection{Retrieval-Based In-Context Learning for Grounding}
The retrieval-based in-context learning (ICL) is illustrated in Fig. \ref{fig:method}(b).
Given an image query $I$ and a sentence (or phrase) query $s$, the instruction-tuned model $LVLM$ predicts a set of bounding boxes $\hat{b}$ on $I$ corresponding to $s$, i.e., $\hat{b} = LVLM(I, s)$.
We use ICL to adapt $LVLM$ to new tasks without fine-tuning by leveraging a small set of demonstrations at inference time.
The $LVLM$ predicts $\hat{b}$ conditioned on the set of demonstrations $C$, i.e., $\hat{b} = LVLM(I, s \mid C)$, where $C = \{(I_i, s_i, b_i)\}_{i=1}^k$ denotes a set of $k$ image-sentence-box triples.
Following retrieval-based approaches \cite{yang2022empirical}, we construct $C$ by selecting the top-$k$ most relevant triplets from the candidate pool.
Given a candidate pool $P$ and an image encoder $E_I$, we retrieve relevant demonstrations by computing $\cos\big(E_I(I), E_I(I_p)\big)$ for all $I_p \in P$, where $\cos(\cdot, \cdot)$ denotes cosine similarity.
Since ICL performance relies on the quality of retrieved demonstrations, the fine-grained representations captured by $E_I$ play a crucial role in effective grounding, underscoring the value of fine-grained representation learning in our framework.

\subsubsection{Discussion}
In phrase grounding, inter-observer variability leads to inconsistent bounding box annotations, posing challenges for model training and evaluation \cite{de2025padchest}.
Retrieval-based ICL is particularly effective in this setting, as it mitigates annotation inconsistency at inference time without additional fine-tuning.
In contrast, when training is feasible, models can be fine-tuned with $\mathcal{L}_g$, thereby further enhancing performance on downstream tasks.
Across both settings, strong fine-grained representations are essential for achieving high performance, which aligns with our training objective.

\subsubsection{Implementation Details}
We adopted a pretrained SigLIP2-400M \cite{tschannen2025siglip} with an input size of 512 as $E_I$ and $E_T$, and used Gemma-3-270M \cite{team2025gemma} as the $LLM$. Following prior work \cite{lai2024carzero,xiao2025flair}, we randomly sampled 4-8 sentences from radiology reports for contrastive learning due to the 64 token limit of the text encoder. We normalized bounding box coordinates to $[0,1000]$ and converted them into strings to compute autoregressive $\mathcal{L}_g$ and $\mathcal{L}_d$. We applied LoRA \cite{hu2022lora} with rank 16 to the query, key, value, and output layers of $E_I$ and $E_T$, and with rank 4 to the query and value layers of the $LLM$.
We trained the model for 10 epochs with a batch size of 16 using cosine annealing with an initial learning rate of 3e-4.
For retrieval-based ICL, we used instruction-tuned MedGemma-1.5-4B \cite{sellergren2025medgemma} with an input size of 896. Other LVLMs \cite{liu2025gemex,li2025knowledge,chen2024chexagent,bannur2024maira,deperrois2025radvlm} had limited context lengths (e.g., 4096 tokens) and therefore could not support ICL.


\section{Experiments and Results}
\subsubsection{Datasets} \textbf{MIMIC-CXR} is a dataset consisting of 377,110 chest X-rays and 227,835 radiology reports \cite{johnson2019mimic}. We utilized the findings and impression sections annotated by an LLM in \cite{zambrano2025clinically} and excluded non-frontal and low-quality chest X-rays for our experiments using \cite{gaggion2024chexmask}.
For grounding annotations, a total of 394,835 A\textsuperscript{++}-rated text-box pairs from 153,602 studies in MIMIC-Ext \cite{mimic-ext-cxr-qba} were used.
As the dataset provides machine-generated silver-standard ground-truth annotations, we applied a weighted box fusion strategy \cite{solovyev2021weighted} to merge duplicate boxes. Notably, unlike typical phrase grounding datasets for abnormal findings (e.g., PadChest-GR \cite{de2025padchest}), MIMIC-Ext \cite{mimic-ext-cxr-qba} consists of anatomical-level bounding box annotations, providing comparatively larger bounding boxes. The dataset was used to train our model on the official training split, and the model's fine-grained representations were evaluated on 2,432 patient studies from the official test split via retrieval tasks.
\textbf{PadChest-GR} is a dataset consisting of 4,555 chest X-rays paired with grounding reports containing text-box annotations \cite{de2025padchest}. We exclusively used annotations with non-empty bounding boxes in our phrase grounding experiments.
We used 1,238 text-box pairs from 604 chest X-rays in the official test split for evaluation and 4,341 text-box pairs from 2,096 chest X-rays for training (or as the candidate pool).

\begin{table}[t]
\centering
\caption{Retrieval performance on the MIMIC-CXR dataset. R@K denotes Recall at K, where K is the cutoff rank.}
\label{tab:retrieval}
\begin{tabular}{l|ccccc|ccccc}
\hline
\multirow{2}{*}{Model} & \multicolumn{5}{c|}{Image-to-Text} & \multicolumn{5}{c}{Text-to-Image} \\
& R@1 & R@5 & R@10 & R@20 & R@40 & R@1 & R@5 & R@10 & R@20 & R@40 \\
\hline
SigLIP2 \cite{tschannen2025siglip}
& \heatRet{0.08} & \heatRet{0.95} & \heatRet{1.56} & \heatRet{2.59} & \heatRet{4.32}
& \heatRet{0.27} & \heatRet{0.98} & \heatRet{1.60} & \heatRet{2.95} & \heatRet{4.69} \\
BiomedCLIP \cite{zhang2025multimodal}
& \heatRet{0.82} & \heatRet{2.10} & \heatRet{3.78} & \heatRet{7.07} & \heatRet{12.66}
& \heatRet{0.58} & \heatRet{1.77} & \heatRet{3.17} & \heatRet{6.00} & \heatRet{11.02} \\
BMC-CLIP \cite{lozano2025biomedica}
& \heatRet{0.12} & \heatRet{0.53} & \heatRet{1.15} & \heatRet{2.55} & \heatRet{4.44}
& \heatRet{0.05} & \heatRet{0.80} & \heatRet{1.65} & \heatRet{2.63} & \heatRet{4.78} \\
MedSigLIP \cite{sellergren2025medgemma}
& \heatRet{4.61} & \heatRet{14.27} & \heatRet{21.92} & \heatRet{32.69} & \heatRet{44.82}
& \heatRet{2.38} & \heatRet{10.33} & \heatRet{17.47} & \heatRet{26.43} & \heatRet{38.48} \\
CARZero \cite{lai2024carzero}
& \heatRet{11.64} & \heatRet{29.52} & \heatRet{40.17} & \heatRet{51.19} & \heatRet{63.98}
& \heatRet{9.67} & \heatRet{26.02} & \heatRet{36.24} & \heatRet{48.40} & \heatRet{62.37} \\
RadIR \cite{zhang2025radir}
& \heatRet{7.36} & \heatRet{22.62} & \heatRet{32.48} & \heatRet{44.70} & \heatRet{57.44}
& \heatRet{7.11} & \heatRet{22.70} & \heatRet{32.61} & \heatRet{45.44} & \heatRet{59.91} \\
\textbf{Ours}
& \heatRet{13.90} & \heatRet{35.57} & \heatRet{47.29} & \heatRet{60.20} & \heatRet{72.53}
& \heatRet{11.91} & \heatRet{31.30} & \heatRet{42.36} & \heatRet{54.46} & \heatRet{66.71} \\
\hline
\end{tabular}
\end{table}

\subsection{Retrieval Experiments}
We evaluated the fine-grained representations of our model on the MIMIC-CXR dataset using image-to-text and text-to-image retrieval tasks. We split each radiology report into five-sentence sub-reports and evaluated report-level retrieval performance using Recall at K (R@K), where K is the cutoff rank \cite{xiao2025flair}.
We compared our method against several contrastive-based representation learning models, including SigLIP2 \cite{tschannen2025siglip}, BiomedCLIP \cite{zhang2025multimodal}, and BMC-CLIP \cite{lozano2025biomedica}, as well as models trained on MIMIC-CXR, such as MedSigLIP \cite{sellergren2025medgemma}, CARZero \cite{lai2024carzero}, and RadIR \cite{zhang2025radir}.
We loaded the publicly released checkpoints of the comparison methods and evaluated them on the test set.

Table \ref{tab:retrieval} shows the retrieval performance on the MIMIC-CXR dataset. General medical contrastive models achieve lower retrieval performance than models tailored to chest X-ray data, highlighting the advantage of modality-specific approaches. MedSigLIP exhibits lower accuracy than CARZero and RadIR, which can be attributed to its training on medical data alongside a large proportion of non-medical images. Our method outperforms state-of-the-art contrastive models trained on MIMIC-CXR in both retrieval directions, demonstrating its effectiveness in learning fine-grained representations.

\subsection{Phrase Grounding Experiments}
We conducted phrase grounding experiments on the PadChest-GR dataset to evaluate fine-grained representations in external validation using retrieval-based ICL and downstream performance in internal validation using pretrained parameters.
In external validation, PadChest-GR was used exclusively for evaluation, whereas in internal validation, it was used for both model training and evaluation.
We reported the localization (Ro/L) and shape-matching (Ro/S) scores of robust detection outcome \cite{meissen2024robust}, as well as precision (P@0.5), recall (R@0.5), and F1 score (F@0.5) at an intersection over union threshold of 0.5 (rather than average precision, which relies on confidence scores) \cite{jiang2025detect}. All models were evaluated under a single-class setting to facilitate comparison with each other.

\begin{table}[t]
\centering
\caption{Phrase grounding performance on the PadChest-GR dataset for external validation using retrieval-based ICL with MedGemma and $k$ demonstrations.}
\label{tab:ground_icl}
\begin{tabular}{l|cc|ccc|cc|ccc}
\hline
\multirow{2}{*}{Model} 
& \multicolumn{5}{c|}{$k$=20}
& \multicolumn{5}{c}{$k$=40} \\
& Ro/L & \multicolumn{1}{c}{Ro/S} & F@0.5 & P@0.5 & R@0.5
& Ro/L & \multicolumn{1}{c}{Ro/S} & F@0.5 & P@0.5 & R@0.5 \\
\hline
SigLIP2 \cite{tschannen2025siglip}
& \heatRo{52.50} & 42.14 & \heatDet{17.75} & 18.61 & 16.96
& \heatRo{56.87} & 44.70 & \heatDet{21.47} & 22.53 & 20.51 \\
BiomedCLIP \cite{zhang2025multimodal}
& \heatRo{56.19} & 42.39 & \heatDet{18.43} & 19.57 & 17.42
& \heatRo{60.14} & 45.40 & \heatDet{22.06} & 22.94 & 21.24 \\
BMC-CLIP \cite{lozano2025biomedica}
& \heatRo{53.24} & 42.20 & \heatDet{17.15} & 18.16 & 16.24
& \heatRo{56.71} & 44.66 & \heatDet{20.92} & 21.95 & 19.99 \\
MedSigLIP \cite{sellergren2025medgemma}
& \heatRo{58.79} & 44.69 & \heatDet{21.77} & 22.62 & 20.97
& \heatRo{61.63} & 46.89 & \heatDet{24.51} & 25.12 & 23.93 \\
RadIR \cite{zhang2025radir}
& \heatRo{56.03} & 42.13 & \heatDet{17.50} & 18.54 & 16.57
& \heatRo{56.34} & 41.72 & \heatDet{17.04} & 17.77 & 16.37 \\
\textbf{Ours}
& \heatRo{59.10} & 44.49 & \heatDet{22.48} & 23.40 & 21.63
& \heatRo{63.55} & 48.00 & \heatDet{25.34} & 25.98 & 24.72 \\
\hline
\end{tabular}
\end{table}

\begin{figure}[t]
\centering
\begin{minipage}[t]{0.59\textwidth}
\centering
\captionof{table}{Phrase grounding performance on the PadChest-GR dataset for external validation.}
\label{tab:ground_ext}
\begin{tabular}{l|cc|ccc}
\hline
Model & Ro/L & Ro/S & F@0.5 & P@0.5 & R@0.5 \\
\hline
MedRPG \cite{chen2023medical}
& 31.44 & 24.00 & 2.10 & 2.34 & 1.91 \\
ChEX \cite{muller2024chex}
& 39.73 & 30.98 & 3.09 & 2.97 & 3.22 \\
GEMeX \cite{liu2025gemex}
& 27.93 & 16.10 & 9.36 & 15.15 & 6.77 \\
MedGemma 
& 38.66 & 27.01 & 2.03 & 2.27 & 1.84 \\
K2Sight \cite{li2025knowledge}
& 49.81 & 25.61 & 8.68 & 9.46 & 8.02 \\
\textbf{Ours} \w ICL
& \textbf{63.55} & \textbf{48.00} & \textbf{25.34} & \textbf{25.98} & \textbf{24.72} \\
\hline
\end{tabular}
\end{minipage}
\begin{minipage}[t]{0.40\textwidth}
\centering
\captionsetup{skip=0pt}
\captionof{figure}{Qualitative results with ground truth (dashed) and predictions (solid).}
\includegraphics[width=\textwidth]{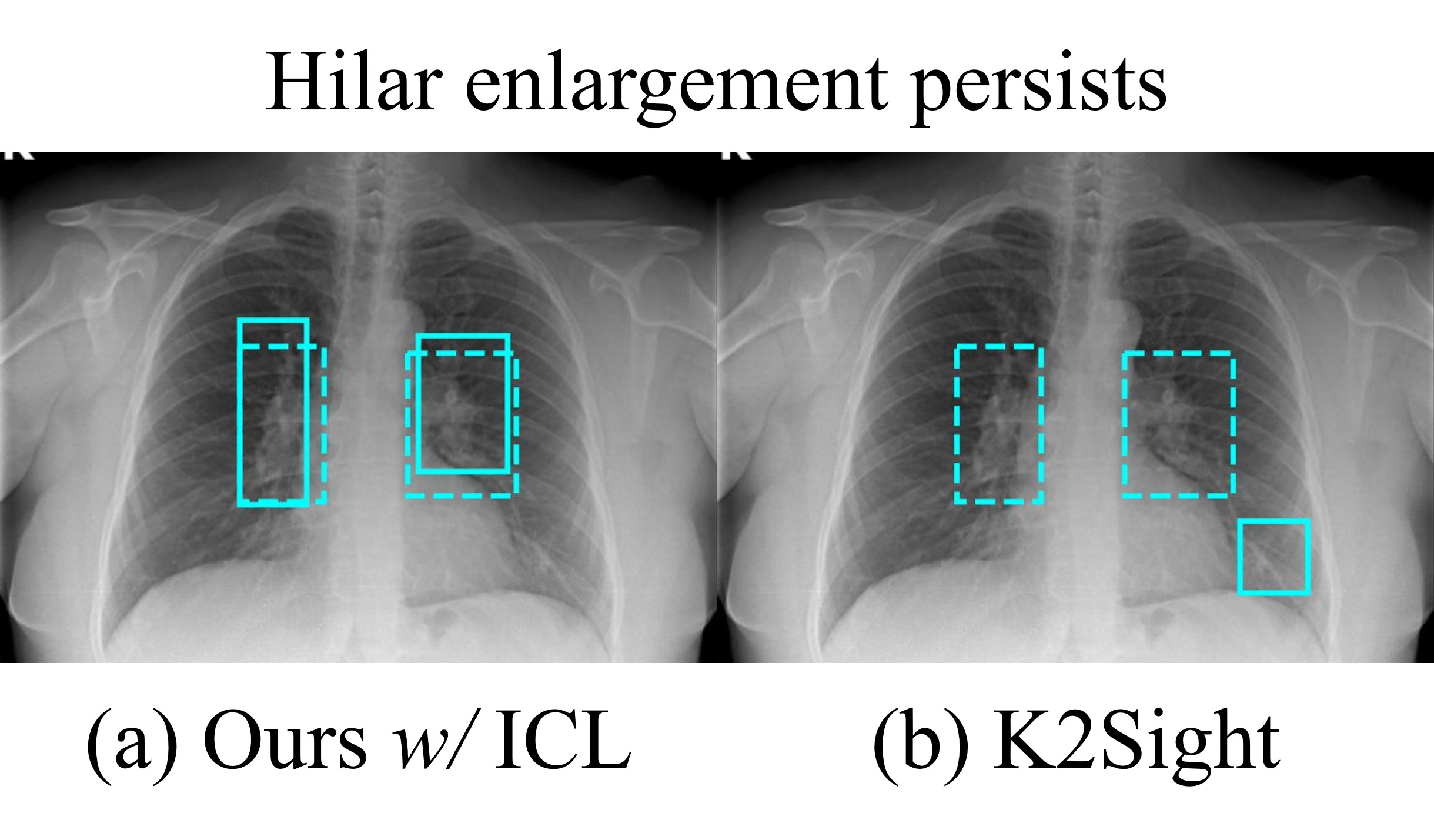}
\label{fig:qual_ext}
\end{minipage}
\end{figure}

\subsubsection{External Validation}
We evaluated the phrase grounding performance of contrastive models \cite{tschannen2025siglip,zhang2025multimodal,lozano2025biomedica,sellergren2025medgemma,zhang2025radir} using retrieval-based ICL with MedGemma \cite{sellergren2025medgemma}. We assessed performance across different settings by varying the number of demonstrations ($k$=20 and $k$=40). CARZero \cite{lai2024carzero} was excluded as it only supports cross-modal similarity. Additionally, we compared our method (\w ICL) against existing chest X-ray grounding models, including MedRPG \cite{chen2023medical}, ChEX \cite{muller2024chex}, GEMeX \cite{liu2025gemex}, K2Sight \cite{li2025knowledge}, and MedGemma \cite{sellergren2025medgemma}.

Table \ref{tab:ground_icl} shows the phrase grounding performance on the PadChest-GR dataset for external validation using retrieval-based ICL. Compared to MedGemma's performance without ICL in Table \ref{tab:ground_ext}, incorporating retrieval-based ICL with contrastive models leads to a significant performance improvement, and additional gains are observed as the number of demonstrations increases. Among these models, our method achieves the best performance, demonstrating its effectiveness in capturing fine-grained representations for real-world settings.

Table \ref{tab:ground_ext} shows the phrase grounding performance of chest X-ray grounding models. Although MedRPG and ChEX are designed for grounding abnormal findings, they exhibit limited performance due to inter-observer variability. Similarly, models trained on anatomical-level MIMIC-CXR annotations, including GEMeX and MedGemma (\wo ICL), show degraded performance. K2Sight, designed to improve reliability on PadChest-GR through interpretable visual attribute decomposition, similarly shows limited performance (see Fig. \ref{fig:qual_ext}(b)). Overall, our method with retrieval-based ICL ($k$=40) outperforms all baselines, highlighting its potential for real-world chest X-ray grounding.

\begin{figure}[t]
\centering
\begin{minipage}[t]{0.59\textwidth}
\centering
\captionof{table}{Phrase grounding performance on the PadChest-GR dataset for internal validation.}
\label{tab:ground_int}
\begin{tabular}{l|cc|ccc}
\hline
Model & Ro/L & Ro/S & F@0.5 & P@0.5 & R@0.5 \\
\hline
CheXagent \cite{chen2024chexagent}
& 27.19 & 14.93 & 11.30 & 20.81 & 7.76 \\
MAIRA-2 \cite{bannur2024maira}
& 53.57 & 35.87 & 32.93 & \textbf{40.09} & 27.94 \\
RadVLM \cite{deperrois2025radvlm}*
& 70.21 & 44.85 & 29.27 & 27.70 & 31.03 \\
FT \wo Ours
& 65.54 & 46.60 & 28.39 & 27.81 & 28.99 \\
FT \textbf{\w Ours}
& \textbf{70.42} & \textbf{47.97} & \textbf{33.44} & 33.76 & \textbf{33.14} \\
\hline
\multicolumn{6}{l}{\scriptsize * indicates test data seen during training.} \\
\end{tabular}
\end{minipage}
\begin{minipage}[t]{0.40\textwidth}
\centering
\captionsetup{skip=0pt}
\captionof{figure}{Qualitative results with ground truth (dashed) and predictions (solid).}
\includegraphics[width=\textwidth]{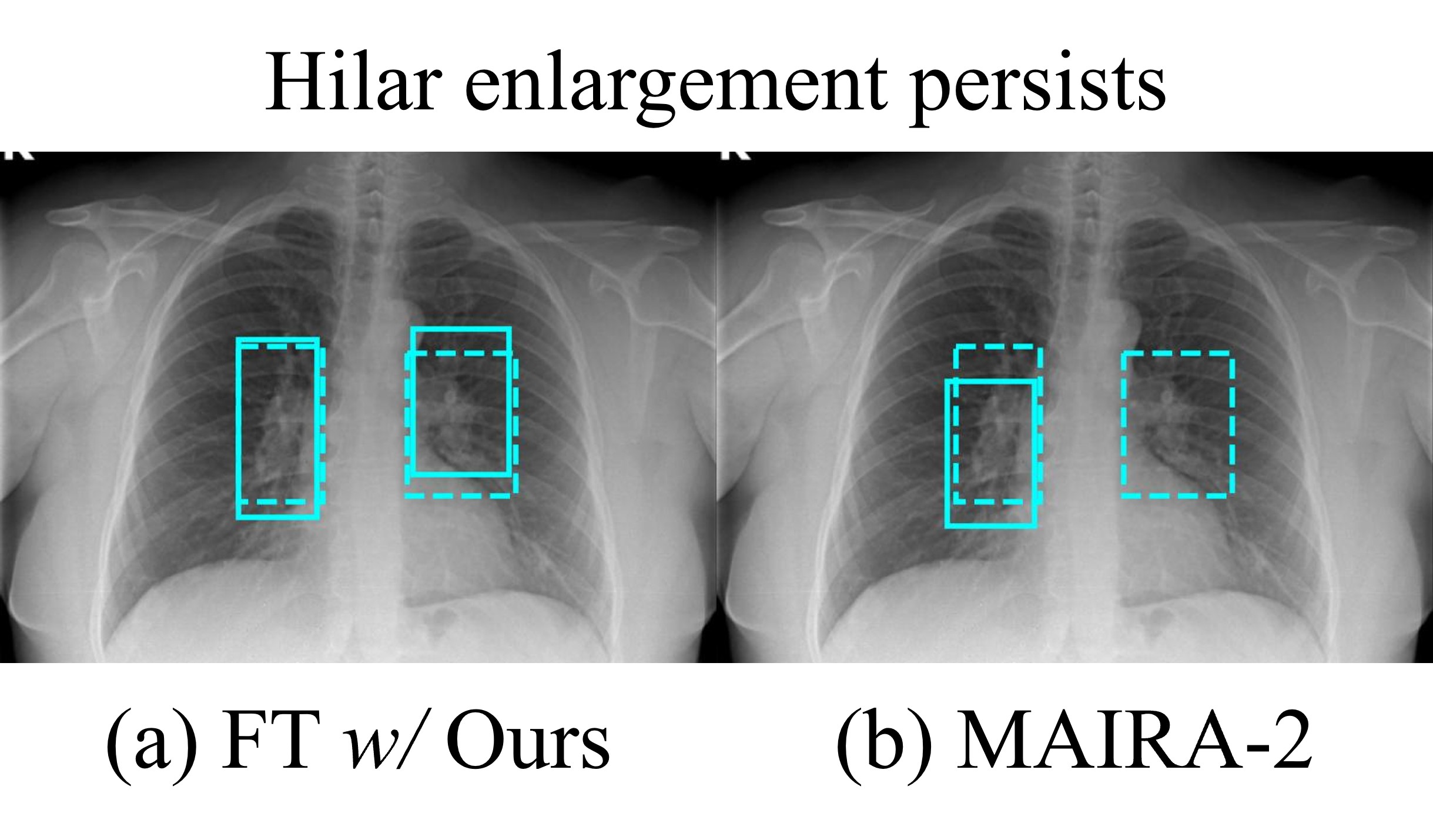}
\label{fig:qual_int}
\end{minipage}
\end{figure}

\subsubsection{Internal Validation}
We compared our method (FT \w Ours) against chest X-ray grounding LVLMs trained on PadChest-GR, including CheXagent \cite{chen2024chexagent}, MAIRA-2 \cite{bannur2024maira}, and RadVLM \cite{deperrois2025radvlm}.
The performance of RadVLM may be inflated, as the model was trained on the full PadChest-GR dataset. Additionally, we evaluated a variant of our model without MIMIC-CXR pre-training (FT \wo Ours) to quantify its impact.

Table \ref{tab:ground_int} shows the phrase grounding performance on the PadChest-GR dataset for internal validation. MAIRA-2 shows relatively higher P@0.5 but lower R@0.5 and Ro/L performance, likely due to its cautious bounding box predictions (see Fig. \ref{fig:qual_int}(b)). FT \wo Ours outperforms MAIRA-2 on the Ro/L metric but achieves lower performance on F@0.5. By leveraging pretrained parameters, FT \w Ours improves F@0.5 by 5\% compared to FT \wo Ours, surpassing MAIRA-2 on both Ro/L and F@0.5 metrics. Although the performance gains in F@0.5 are marginal compared to MAIRA-2, our model is more parameter-efficient, using 270M parameters, whereas MAIRA-2 uses 7B. In summary, the proposed objective facilitates the learning of pretrained parameters that significantly improve downstream fine-grained grounding performance.

\begin{figure}[t]
\centering
\begin{subfigure}{0.19\linewidth}
\includegraphics[width=\linewidth]{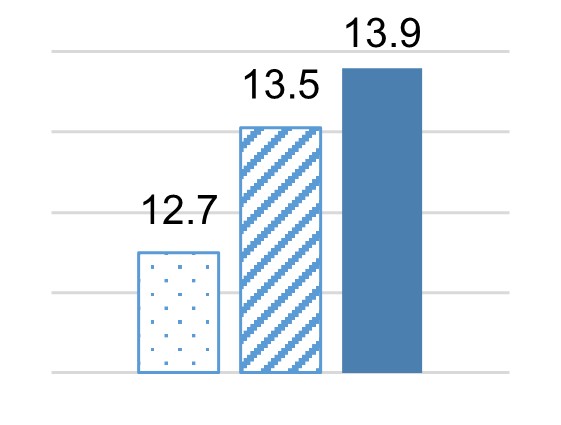}
\caption{R@1}
\end{subfigure}
\begin{subfigure}{0.19\linewidth}
\includegraphics[width=\linewidth]{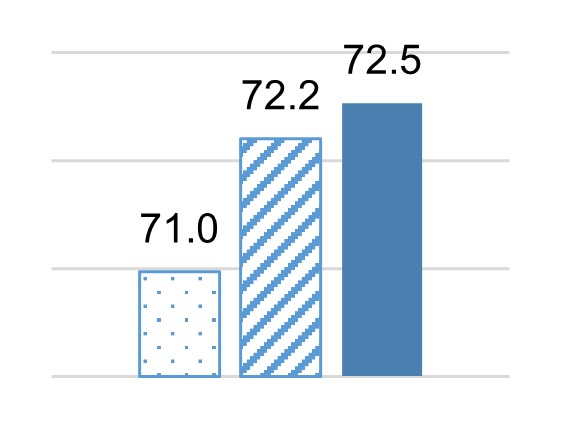}
\caption{R@40}
\end{subfigure}
\begin{subfigure}{0.19\linewidth}
\includegraphics[width=\linewidth]{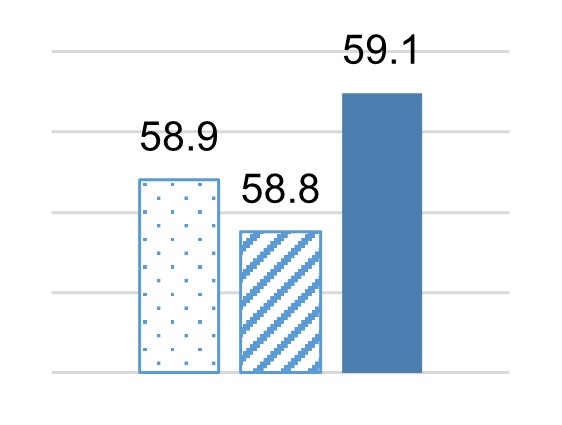}
\caption{Ro/L}
\end{subfigure}
\begin{subfigure}{0.19\linewidth}
\includegraphics[width=\linewidth]{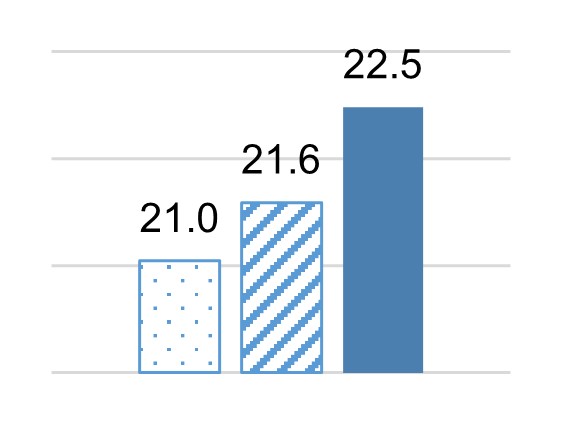}
\caption{F@0.5}
\end{subfigure}
\begin{subfigure}{0.19\linewidth}
\includegraphics[width=\linewidth]{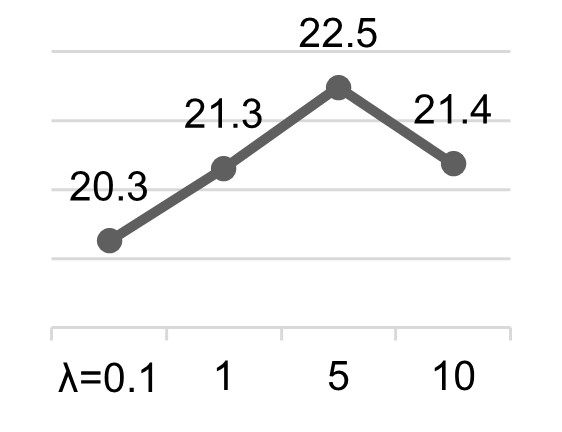}
\caption{F@0.5 ($\lambda$)}
\end{subfigure}
\caption{(a)-(d) Ablation results of our method using all losses (filled), without $\mathcal{L}_{g,d}$ (hatched), and without both $\mathcal{L}_c$ and $\mathcal{L}_{g,d}$ (dotted). (e) Ablation results for different $\lambda$ values (line graph).}
\label{fig:abl}
\end{figure}

\subsubsection{Ablation Studies}
We performed ablation studies on the retrieval task in MIMIC-CXR and the grounding task in PadChest-GR. We evaluated variants of our method without $\mathcal{L}_{g,d}$ and without both $\mathcal{L}_c$ and $\mathcal{L}_{g,d}$. Additionally, we evaluated our method using different values of $\lambda$. Since the dataset contains samples without bounding boxes, we did not consider the variant without $\mathcal{L}_c$. We reported (a) R@1, (b) R@40, (c) Ro/L ($k$=20), and (d)-(e) F@0.5 ($k$=20).

Fig. \ref{fig:abl}(a)-(d) show the ablation results on both tasks. These results suggest that incorporating the location-aware loss is beneficial for fine-grained representation learning. Fig. \ref{fig:abl}(e) shows performance with varying $\lambda$. These results indicate that $\lambda$=5 performs best and that assigning a moderately higher weight to the proposed components benefits training. In summary, each loss plays an important role in fine-grained representation learning.
\section{Conclusion}
We propose LoFi to address the limitations of existing medical contrastive models and LVLMs in chest X-ray retrieval and phrase grounding. By jointly optimizing sigmoid, captioning, and location-aware captioning losses with a lightweight LLM, our model facilitates learning from long-form text and region-level supervision, enhancing fine-grained representations. Extensive experiments demonstrate that our method consistently achieves superior performance in retrieval on MIMIC-CXR and in phrase grounding on PadChest-GR under both internal and external validation settings. 
The computational overhead introduced by ICL may limit the scalability of the proposed framework, highlighting the need for future research to alleviate this bottleneck and enable more efficient deployment.

\subsubsection{Acknowledgments}
The research is supported, in part, by the NSERC Discovery Grant RGPIN-2022-05316, NSERC Alliance Grant ALLRP 602633-24, Tri-Agency Canada; Canada CIFAR AI Chair Awards, and Canada Research Chair Fellowship; Google Gemini Research Awards, IITP grant, the Ministry of Science and ICT (RS-2024-00445087, RS-2025-25464461), funded by the Korea government (MSIT); the National Research Foundation of Korea (NRF) grant funded by the Korea government (MSIT) (RS-2025-00515536).

%
\bibliographystyle{splncs04}
\bibliography{refs}

\end{document}